\documentclass[10pt,twocolumn,letterpaper]{article}

\usepackage{cvpr,times,graphicx,amsmath,amssymb,algorithm,algorithmic,authblk}
\usepackage[pagebackref=true,colorlinks=true,bookmarks=false]{hyperref}

\newcommand{\app}{\raise.17ex\hbox{$\scriptstyle\sim$}}

\newcommand{\alg}[1]{\textsc{#1}}
\newcommand{\E}{\mathcal{E}}
\cvprfinalcopy
\def\cvprPaperID{472}
\ifcvprfinal\fi

\begin{document}

\title{Unsupervised Learning of Edges\vspace{-2ex}}
\author[1,2]{Yin Li}
\author[1]{Manohar Paluri}
\author[2]{James M. Rehg}
\author[1]{Piotr Doll\'ar}
\affil[1]{Facebook AI Research (FAIR)}
\affil[2]{Georgia Institute of Technology\vspace{-2.5ex}}

\maketitle

\begin{abstract}
Data-driven approaches for edge detection have proven effective and achieve top results on modern benchmarks. However, all current data-driven edge detectors require manual supervision for training in the form of hand-labeled region segments or object boundaries. Specifically, human annotators mark semantically meaningful edges which are subsequently used for training. Is this form of strong, high-level supervision actually necessary to learn to accurately detect edges? In this work we present a simple yet effective approach for training edge detectors without human supervision. To this end we utilize motion, and more specifically, the only input to our method is noisy semi-dense matches between frames. We begin with only a rudimentary knowledge of edges (in the form of image gradients), and alternate between improving motion estimation and edge detection in turn. Using a large corpus of video data, we show that edge detectors trained using our unsupervised scheme approach the performance of the same methods trained with full supervision (within 3-5\%). Finally, we show that when using a deep network for the edge detector, our approach provides a novel pre-training scheme for object detection.
\end{abstract}

\begin{figure}\centering
\includegraphics[width=.49\textwidth]{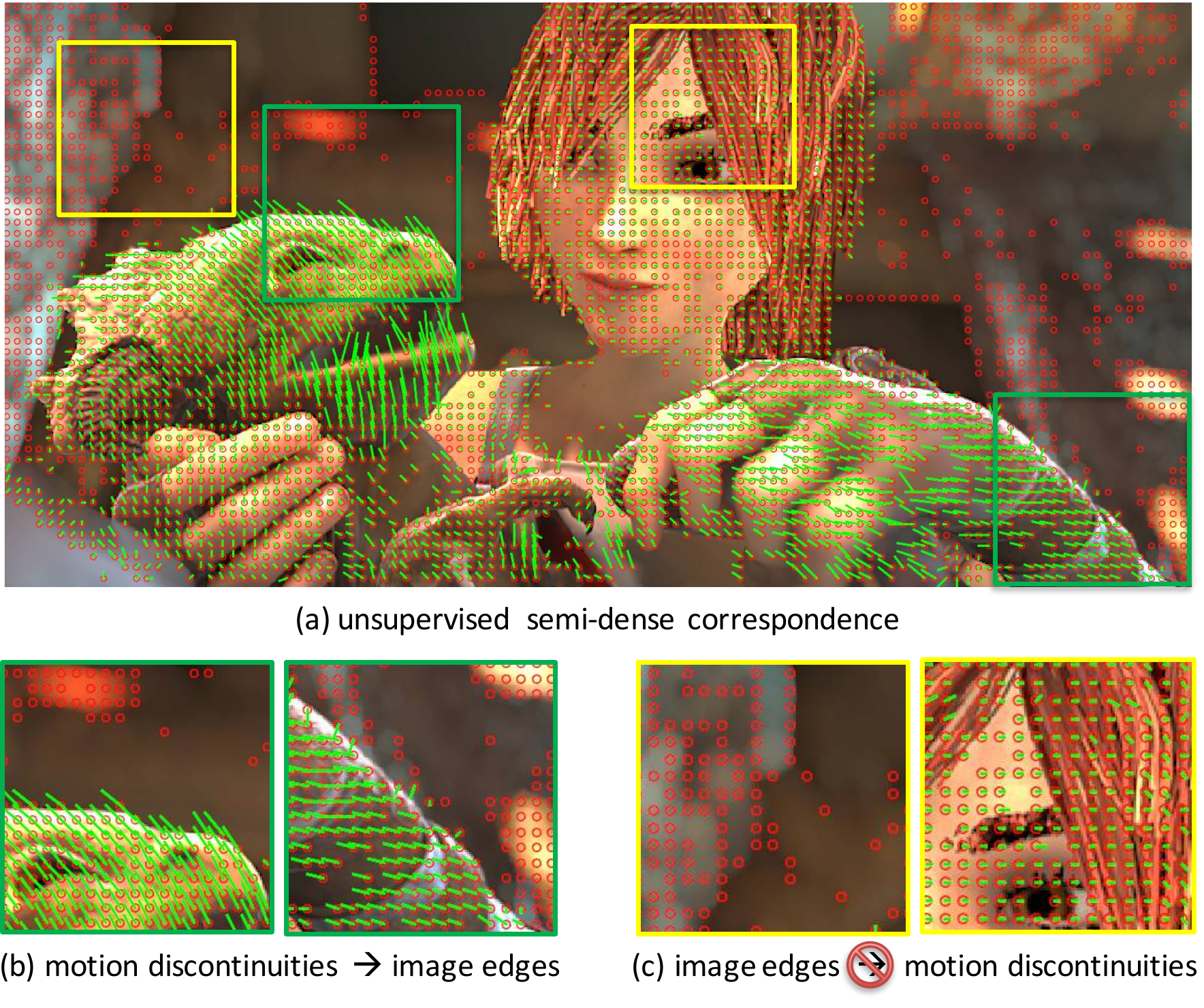}
\caption{Our goal is to train an edge detector given only semi-dense matches between frames (a). While motion discontinuities imply the presence of image edges (b), the converse is not necessarily true as distinct image regions may undergo similar motion (c). In this work we exploit the sparsity of edges to overcome the latter difficulty.  We show that the signal obtained from matches computed over a large corpus of video data is sufficient to train top-performing edge detectors.}
\label{fig:teaser}
\end{figure}

\section{Introduction}

The human visual system can easily identify perceptually salient edges in an image. Endowing machine vision systems with similar capabilities is of interest as edges are useful for diverse tasks such as optical flow~\cite{revaud2015epic}, object detection~\cite{UllmanPAMI91,Ferrari2008PAMI}, and object proposals~\cite{Uijlings13IJCV,Zitnick2014ECCV,Arbelaez2014CVPR}. However, edge detection has proven challenging. Early approaches~\cite{Fram1975IEEE, Canny1986PAMI, Freeman91PAMI} relied on low-level cues such as brightness and color gradients. Reasoning about texture~\cite{Martin2004PAMI} markedly improved results, nevertheless, accuracy still substantially lagged human performance.

The introduction of the BSDS dataset~\cite{Arbelaez2011PAMI}, composed of human annotated region boundaries, laid the foundations for a fundamental shift in edge detection. Rather than rely on complex hand-designed features, Doll\'ar et al.~\cite{Dollar2006CVPR} proposed a data-driven, supervised approach for learning to detect edges. Modern edge detectors have built on this idea and substantially pushed the state-of-the-art forward using more sophisticated learning paradigms \cite{Ren2012NIPS, Lim2013CVPR, Dollar2015fast, Xie2015hed}.

However, existing data-driven methods require strong supervision for training. Specifically, in datasets such as BSDS~\cite{Arbelaez2011PAMI}, human annotators use their knowledge of scene structure and object presence to mark semantically meaningful edges.\footnote{Human annotation of edge structure in local patches (without context) is quite noisy and is matched by machine vision approaches. Humans excel when given context and the ability to reason about object presence~\cite{Zitnick2012CVPR}.} Moreover, recent edge detectors use Image\-Net pre-training~\cite{bertasuis2015high,Xie2015hed}. In this paper, we explore whether this is necessary: \emph{Is object-level supervision indispensable for edge detection? Moreover, can edge detectors be trained entirely without human supervision?}

\begin{figure*}\centering
\includegraphics[width=.95\textwidth]{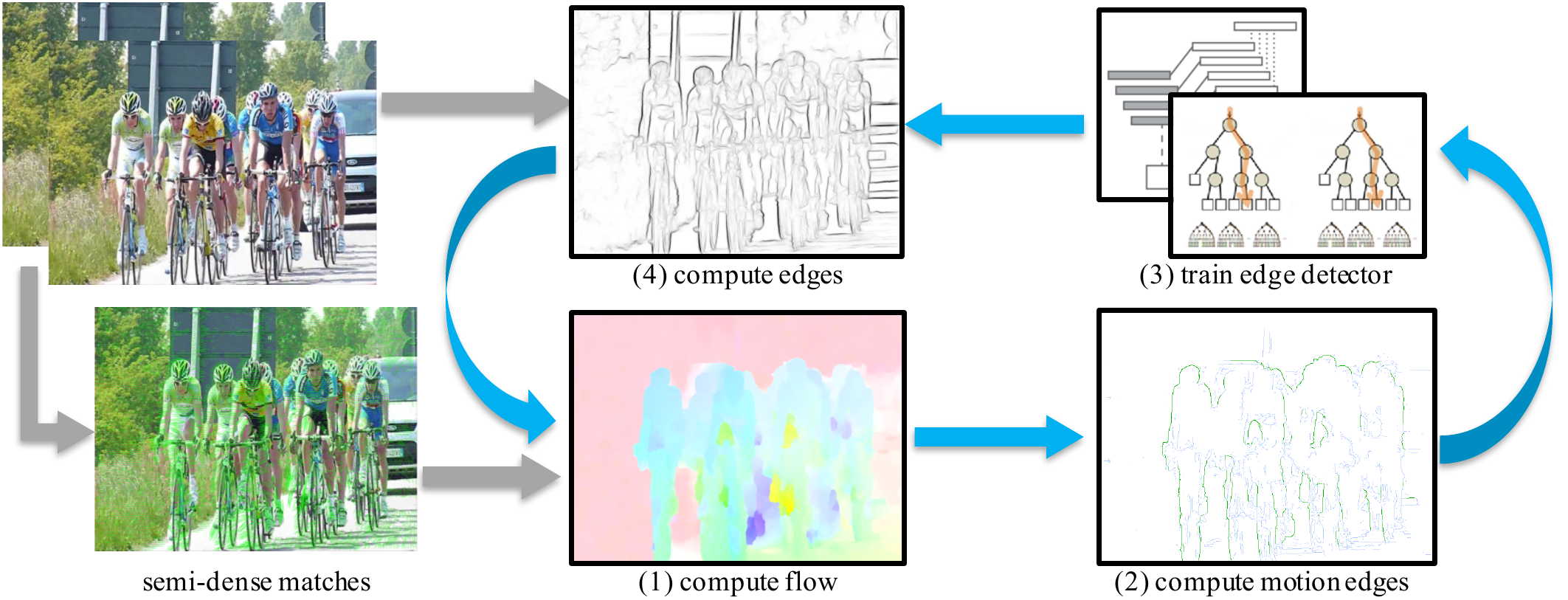}
\caption{The only input to our approach is semi-dense matching results from~\cite{weinzaepfel2013dm}. During training we alternate between: (1)~computing flow based on the matches and edge maps (initialized to simple gradients), (2) computing motion edges from the flow fields (green: positive edge samples; blue: discarded motion edges), (3) training an edge detector using the motion edges as supervision, and (4) recomputing image edges using the new detector. The process is iterated on a large corpus of videos leading to increasingly accurate flow and edges.}
\label{fig:overview}
\end{figure*}

We propose to train edge detectors using motion in place of human supervision. Motion edges are a subset of image edges, see Figure~\ref{fig:teaser}. Therefore motion edges can be used to harvest positive training samples. On the other hand, locations away from motion edges may also contain image edges. Fortunately, as edges are sparse, simply sampling such locations at random can provide good negative training data with few false negatives. Thus, assuming accurate motion estimates, we can potentially harvest unlimited training data for edge detection.

While it would be tempting to assume access to accurate motion estimates, this is arguably an unreasonably strong requirement. Indeed, optical flow and edge detection are tightly coupled. Recently, Revaud et al.~proposed EpicFlow~\cite{revaud2015epic}: given an accurate edge map~\cite{Dollar2015fast} and semi-dense matches between frames~\cite{weinzaepfel2013dm}, EpicFlow generates a dense edge-respecting interpolation of the matches. The result is a state-of-the-art optical flow estimate.

This motivates our approach. We begin with only semi-dense matches between frames~\cite{weinzaepfel2013dm} and a rudimentary knowledge of edges (simple image gradients). We then repeatedly alternate between computing flow based on the matches and most recent edge maps and retraining an edge detector based on signal obtained from the flow fields. Specifically, at each iteration, we first estimate dense flow fields by interpolating the matching results using the edge maps obtained from the previous iteration. Given a large corpus of videos, we next harvest highly confident motion edges as positives and randomly sample negatives, and use this data to train an improved edge detector. The process is iterated leading to increasingly accurate flow and edges. An overview of our method is shown in Figure~\ref{fig:overview}.

We perform experiments with the Structured Edge (SE)~\cite{Dollar2015fast} and Holistic Edge (HE)~\cite{Xie2015hed} detectors. SE is based on structured forests, HE on deep networks; SE is faster, HE more accurate. Both detectors achieve state-of-the-art results. The main result of our paper is that both methods, trained using our unsupervised scheme, approach the level of performance of fully supervised training.

Finally, we demonstrate that our approach can serve as a novel unsupervised pre-training scheme for deep networks~\cite{wang2015unsupervised,doersch2015unsupervised}. Specifically, we show that when fine-tuning a network for object detection~\cite{everingham2014pascal}, starting with the weights learned for edge detection improves performance over starting with a network with randomly initialized weights. While the gains are modest, we believe this is a promising direction for future exploration.

\section{Related Work}

\textbf{Edge Detection:} Early edge detectors were manually designed to use image gradients~\cite{Fram1975IEEE, Canny1986PAMI, Freeman91PAMI} and later texture gradients~\cite{Arbelaez2011PAMI}. Of more relevance to this work are edge detectors trained in a data-driven manner. Since the work of~\cite{Dollar2006CVPR}, which formulated edge detection as a binary classification problem, progressively more powerful learning paradigms have been employed, including multi-class classification~\cite{Lim2013CVPR}, feature learning~\cite{Ren2012NIPS}, regression~\cite{sironi2014multiscale}, structured prediction~\cite{Dollar2015fast}, and deep learning~\cite{Kivinen2014AISTATS, bertasuis2015high, Xie2015hed}. Recently, Weinzaepfel et al.~\cite{weinzaepfel2015learning} extended~\cite{Dollar2015fast} to motion edge estimation. These methods all require strong supervision for training. In this work we explore whether unsupervised learning can be used instead (and as discussed select \cite{Dollar2015fast,Xie2015hed} for our experiments).

\textbf{Optical Flow:} The estimation of optical flow is a classic problem in computer vision~\cite{horn1981determining, Lucas1981iterative}. A full overview is outside of our scope, instead, our work is most closely related to methods that leverage sparse matches or image edges for flow estimation~\cite{brox2011large, weinzaepfel2013dm, revaud2015epic}. In particular, as in~\cite{revaud2015epic}, we use edge-respecting sparse-to-dense interpolation of matches to obtain dense motion estimates. Our focus, however, is not on optical flow estimation, instead, we exploit the tight coupling between edge and flow estimation to train edge detectors without human supervision.

\textbf{Perceptual Grouping using Motion:} Motion plays a key role for grouping and object recognition in the human visual system~\cite{koffka2013principles}. In particular, Ostrovsky et al.~\cite{ostrovsky2009visual} studied the visual skills of individuals recovering from congenital blindness and showed that motion cues were essential to help facilitate the development of object grouping and representation. Our work is inspired by these findings: we aim to learn an edge detector using motion cues.

\textbf{Learning from Video:} There is an emerging interest for learning visual representations using video as a supervisory signal, for example by enforcing that neighboring frames have a similar representation~\cite{mobahi2009deep}, learning latent representations for successive frames~\cite{taylor2010convolutional}, or learning to predict missing or future frames~\cite{ranzato2014video, srivastava2015unsupervised}. Instead of simply enforcing various constraints on successive video frames, Wang and Gupta~\cite{wang2015unsupervised} utilized object tracking and enforce that tracked patches in a video should have a similar visual representation. The resulting network generalizes well to surface normal estimation and object detection. As we will demonstrate, our approach can also serve as a novel unsupervised pre-training scheme. However, while in previous approaches the training objective was used as a surrogate to encourage the network to learn a useful representation, our primary goal is to train an edge detector and the learned representation is simply a useful byproduct.

\section{Learning Edges from Video}

We start with a set of low level cues using standard tools in computer vision, including point correspondences and image gradients. We use DeepMatching~\cite{weinzaepfel2013dm} to obtain semi-dense matches $M$ between two consecutive frames $(I,I')$. DeepMatching computes correlations at different locations and scales to generate the matches. Note that contrary to its name, the method involves no deep learning. For the rest of the paper, we fix the matching results $M$.

Our proposed iterative process is described in Figure~\ref{fig:overview} and Algorithm~\ref{fig:algorithm}. We denote the edge detector at iteration $t$ by $\E^t$. For each image $I_j$, we use $E^t_j$ and $G^t_j$ to denote its image edges and motion edges at iteration $t$. We initialize $E^0_j$ to the raw image gradient magnitude of $I_j$, defined as the maximum gradient magnitude over color channels. The gradient magnitude is a simple approximation of image edges, and thus serves as a reasonable starting point.

At each iteration $t$, we use EpicFlow~\cite{revaud2015epic} to generate edge-preserving flow maps $F^t_j$ given matches $M_j$ and previous edges $E^{t-1}_j$. We next apply $\E^{t-1}$ on a colored version of $F^t_j$ to get an estimate of motion edges $G^t_j$. $G^t_j$ is further refined by aligning to superpixel edges. Next, for training our new edge detector $\E^t$, we harvest positives instances using a high threshold on $G^t_j$ and sample random negatives away from any motion edges.

The above process is iterated until convergence (typically 3 to 4 iterations suffice). At each iteration the flow $F^t_j$ and edge maps $E^t_j$ and $G^t_j$ improve. In the following sections we describe the process in additional detail.

\begin{algorithm}[t]
\small
\caption{Iterative Learning Procedure}
\label{fig:algorithm}
\begin{algorithmic}[1]
\REQUIRE Pairs of frames $(I_j,I_j')$, matches $M_j$
\STATE $\E^0$ = gradient magnitude operator, $E^0_j = \E^0(I_j) \enskip\forall j$
\FOR{$t$ in $1...T$}
	\STATE Estimate flow $F^t_j$ using previous edge maps $E^{t-1}_j$ \\
	$\quad F^t_j = EpicFlow(I_j, I_j', M_j, E^{t-1}_j) \enskip\forall j$
	\STATE Detect motion edges $G^t_j$ by applying $\E^{t-1}$ to $F^t_j$ \\
	$\quad G^t_j = \E^{t-1}(FlowToRgb(F^t_j)) \enskip\forall j$
	\STATE Train new edge detector $\E^t$ using motion edges $G^t_j$ \\
	$\quad \E^t = TrainEdgeDetector( \{I_j,G^t_j\})$
	\STATE Apply edge detector $\E^t$ to all frames \\
	$\quad E^t_j = \E^t(I_j) \enskip\forall j$
\ENDFOR
\RETURN $\E^T$ and $\{ E^T_j, F^T_j, G^T_j\}$
\end{algorithmic}
\end{algorithm}

\subsection{Method Details}

\textbf{EpicFlow:} EpicFlow~\cite{revaud2015epic} takes as input an image pair $(I,I')$, semi-dense matches $M$ between the images, and an edge map $E$ for the first frame. It efficiently computes approximate geodesic distance defined by $E$ between all pixels and matched points in $M$. For every pixel, the geodesic distance is used to find its $K$ nearest matches, and the weighted combination of their motion vectors determines the source pixel's motion. A final optimization is performed by a variational energy minimization to produce an edge-preserving flow map with high accuracy. We refer readers to ~\cite{revaud2015epic} for additional details.

\begin{figure}\centering
\includegraphics[width=.49\textwidth]{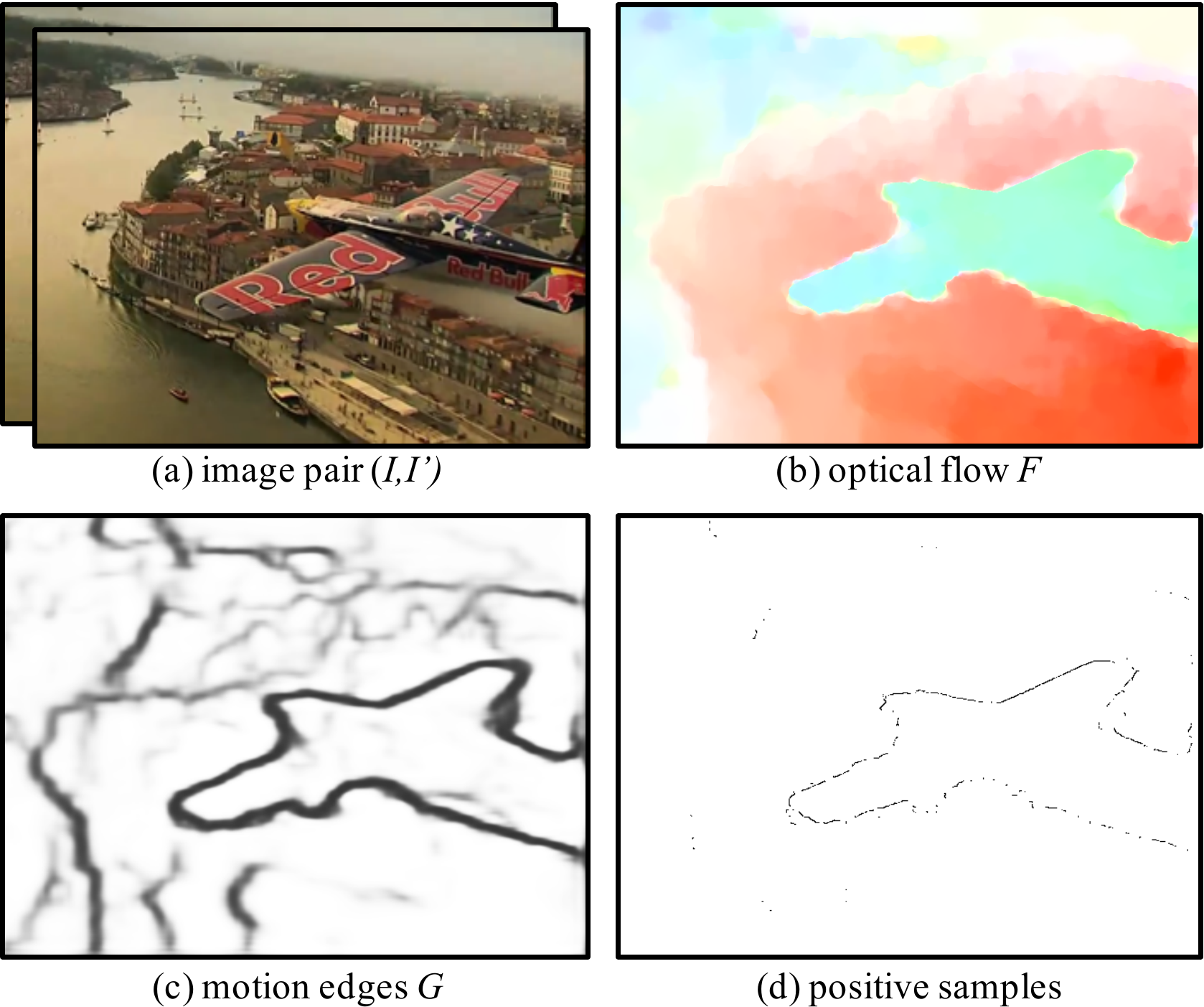}
\caption{Illustration of motion edge detection. (a) Input images. (b) Colorized EpicFlow results $F$ on the input images. (c) Motion edges $G$ computed by applying an edge detector $\E$ to the colorized flow. (d) Motion edges $G$ after alignment, non-maximum suppression, and aggressive thresholding. The aligned motion edge map $G$ serves as a supervisory signal for training an edge detector.}
\label{fig:motion}
\end{figure}

\textbf{Motion Edge Detection:} Detecting motion edges given optical flow estimates can be challenging, see Figure~\ref{fig:motion}. Weinzaepfel et al.~\cite{weinzaepfel2015learning} showed that simply computing gradients over a flow map produces unsatisfactory results and instead proposed a data-driven approach for motion edge detection (for a full review of earlier approaches see~\cite{weinzaepfel2015learning}). In this work we employ a simpler yet surprisingly effective approach. We use an edge detector trained on image edges for motion edge estimation by applying the (image) edge detector to a color-coded flow map. The standard color-coding scheme for optical flow maps 2D flow vectors into a 3D color space by encoding flow orientation via hue and magnitude via saturation~\cite{baker2011database}. Motion edges become clearly visible in this encoding (\ref{fig:motion}b) (we tried other color spaces but HSV worked best). Running an edge detector $\E$ on the colored flow map gives us a simple mechanism for motion edge detection (\ref{fig:motion}c). Moreover, in our iterative scheme, as both our edge detector $\E^{t-1}$ and flow estimate $F^t$ improve with each iteration $t$, so do our resulting estimates of motion edges $G^t=\E^{t-1}(FlowToRgb(F^t))$.

\textbf{Motion Edge Alignment:} Motion edges computed from flow exhibit slight misalignment with their corresponding image edges. We found that this can adversely affect training, especially for HE which produces thick edges. To align the motion edges we apply a simple heuristic: after applying non-maximum suppression and thresholding, we align the motion edges to superpixels detected in the color image. Specifically, we utilize SLIC super-pixels~\cite{achanta2012slic}, which cover over 90\% of all image edges, and match motion and superpixel edge pixels using bipartite matching (also used in BSDS evaluation) with a tolerance of 3 pixels. Matched motion edge pixels are shifted to the superpixel edge locations and unmatched motion edges are discarded.  This refinement, illustrated in Figure \ref{fig:motion}d, helps to filter out edges with weak image gradients and improves localization.

We emphasize that our goal is not to detect all motion edges. A subset with high precision is sufficient for training. Given a large video corpus, high-precision motion edges should provide a dense coverage of image edges. However, due to our alignment procedure our sampling is slightly biased. In particular, motion edges with weak corresponding image edges are often missing. This limitation and its impact on performance is discussed in Section~\ref{sec:exps}.

\textbf{Training $\E$:} The aligned motion edge maps $G^t$ serve as a supervisory signal for training an edge detector $\E^t$. Positives are sampled at locations with high scoring motion edges in $G^t$. Negatives are uniformly sampled from location with motion edges below a small threshold. Note that locations with ambiguous motion edge presence ($G^t$ with intermediate scores) are not considered in training. As we will demonstrate, samples harvested in this manner provide a strong supervisory signal for training $\E$.

\textbf{Video Dataset:} For training, we combine videos from two different datasets: the Video Segmentation Benchmark (VSB)~\cite{galasso2013unified} and the YouTube Object dataset~\cite{Prest2012CVPR}. We use all HD videos ($100+155$) in both datasets. We drop all the annotations for YouTube object dataset. This collection of videos ($\app250$) contains more than $500K$ frames and has sufficient diversity for training an edge detector.

\textbf{Frame Filtering:} Given the vast amount of available data, we apply a simple heuristic to select the most promising frames for motion estimation. We first fit a homography matrix between consecutive frames using ORB descriptor matches~\cite{rublee2011orb} (which are fast to compute). We then remove frames with insufficient matches, very slow motion (max displacement $<$2 pixels), very large motion (average displacement $>$15 pixels), or a global translational motion. These heuristics remove frames where optical flow may be either unreliable or contain few motion edges. For all experiments we used this pruned set of $\app50K$ frames.

\subsection{Edge Detector Details}

We experiment with the Structured Edge (SE)~\cite{Dollar2015fast} and Holistic Edge (HE)~\cite{Xie2015hed} detectors, based on structured forests and deep networks, respectively. SE has been used extensively due to its accuracy and speed, e.g.~for flow estimation~\cite{revaud2015epic} and object proposals~\cite{Zitnick2014ECCV,Arbelaez2014CVPR}. HE is more recent but achieves the best reported results to date. When trained using our unsupervised scheme, both methods approach similar performance as when trained with full supervision.

\textbf{Structured Edges (SE):} SE extracts low-level image features, such as color and gradient channels, to predict edges. The method learns decision trees by using structured labels (patch edge maps) to determine the split function at each node. During testing, each decision tree maps an input patch to a local edge map. The final image edge map is the average of multiple overlapped masks predicted by each tree at each location, leading to a robust and smooth result. We use the same parameters as in~\cite{Dollar2015fast} for training. The forest has 8 trees with maximum depth of 64. Each tree is trained using a random subset (25\%) of $10^6$ patches, with equal number of positives and negatives. During training, we convert a local edge map to a segmentation mask as required by SE by computing connected components in the edge patch. We discard patches that contain edge fragments that do not span the whole patch (which result in a single connected component). During each iteration of training, the forest is learned from scratch. During testing, we run SE over multiple scales with sharpening for best results.

\textbf{Holistic Edges (HE):} HE uses a modified VGG-16 network~\cite{Simonyan14vgg} with skip-layer connections and deep supervision~\cite{lee2014deeply}. Our implementation generally follows~\cite{Xie2015hed}. We remove all fully connected layers and the last pooling layer, resulting in an architecture with 13 conv and 4 max pooling layers. Skip-layers are implemented by attaching linear classifiers ($1\times1$ convolutions) to the last conv layer of each stage, their outputs are averaged to generate the final edge map. In our implementation, we remove the deep supervision (multiple loss functions attached to different layers) as we found that a single loss function has little performance penalty ($.785$ vs $.790$ in ODS score) but is easier to train.

We experimented with both fine-tuning a network pre-trained on ImageNet~\cite{ILSVRC15} and training a network from scratch (random initialization). For fine-tuning, we use the same hyper-parameter as in~\cite{Xie2015hed} with learning rate $1e{-6}$, weight decay $.0002$, momentum $.9$, and batch size $10$. When training from scratch, we add batch normalization~\cite{ioffe2009batch} layers to the end of every conv block. This accelerates training and also improves convergence. We also increase learning rate $(1e{-5})$ and weight decay $(.0005)$ when training from scratch. We train the network for $40$ epochs in each iteration, then reduce learning rate by half. Unlike for SE, we can reuse the network from previous iterations as the starting point for each subsequent iteration.

The somewhat noisy labels, in particular missing positive labels, prove to be challenging for training HE. The issue is partially alleviated by discarding ambiguous samples during back propagation. Furthermore, unlike in~\cite{Xie2015hed}, we randomly select negative samples ($40\times$ as many negatives as positives) and discard negatives with highest loss (following the same motivation as in~\cite{szegedy2014scalable}). Without these steps for dealing with noisy labels convergence is poor.

\section{Experiments and Results}\label{sec:exps}

Our method produces motion edges $G^t$, image edges $E^t$, and optical flow $F^t$ at each iteration $t$. We provide an extensive benchmark for each task tested with two different edge detectors (\alg{se} and \alg{he}). Our main result is that the image edge detectors, trained using videos only, achieve comparable results as when trained with full supervision. As a byproduct of our approach, we also generate competitive optical flow and motion edge results. Finally, we show that pre-training networks using video improves their performance on object detection over training from scratch.

\subsection{Motion Edge Detection}

\begin{table}\centering\small
\begin{tabular}{l|ccc|c}
Method & ODS & OIS & AP & P20 \\
 \hline
 \alg{human} & .63 & .63 & - & -\\
 \alg{se-image} & .45 & .48 & .33 & .39\\
 \alg{he-image} & .47 & .52 & .35 & .49\\
 \alg{epicflow} & .39 & .47 & .33 & .55\\
 \alg{galasso}~\cite{galasso2013unified}\ & .34 & .43 & .23 & .34 \\
 \hline
 \alg{weinzaepfel}~\cite{weinzaepfel2015learning} & .53 & .55 & .37 & \bf.71\\
 \alg{se-video} & .44 & .48 & .34 & \bf.67\\
 \alg{he-video} & .45 & .47 & .32 & \bf.66\\
\end{tabular}\vspace{2mm}
\caption{Motion edge results on the VSB benchmark. See text.}
\label{table:motion}
\end{table}

While our focus is not on motion edge detection, identifying motion edges reliably is important as motion edges serve as our only source of supervision. Thus our first experiment is to benchmark motion edges.

We use the Video Segmentation Benchmark (VSB)~\cite{galasso2013unified} which has annotated ground truth motion edges every $20$ frames. We report results on the $282$ annotated frames in the test set (we remove frames without motion edges and the final frame of each video as \cite{weinzaepfel2015learning} requires $3$ frames). We evaluate using three standard metrics~\cite{Arbelaez2011PAMI}: fixed contour threshold (ODS), per-image best threshold (OIS), and average precision (AP). As we are concerned about the high precision regime, we introduce an additional measure: precision at $20\%$ recall (P20). Non-maximum suppression is applied to all motion edges prior to evaluation.

In Table~\ref{table:motion} we report results of four baselines and the motion edges $G^T$ obtained from the final iteration of our approach (\alg{se/he-video}). The baselines include: image edges (\textsc{se/he-image}), gradient magnitude of optical flow (\alg{epicflow}), a method which combines superpixel segmentation with motion cues (\alg{Galasso}~\cite{galasso2013unified}), and a recent data-driven supervised approach (\alg{Weinzaepfel}~\cite{weinzaepfel2015learning}).

Our method, albeit simple, has a precision $.66 \app .67$ at $20\%$ recall, only slightly worse than~\cite{weinzaepfel2015learning}, even though it was not trained for motion edge detection. It substantially outperforms all other baselines in the high precision regime. While our goal is not motion edge detection per-se, this result is important as it enables us to obtain high quality positive samples for training an image edge detector.

\subsection{Image Edge Detection}

We next investigate edge detection performance. Results are reported on the Berkeley Segmentation Dataset and Benchmark (BSDS)~\cite{Martin2004PAMI,Arbelaez2011PAMI}, composed of 200 train, 100 validation, and 200 test images. Each image is annotated with ground truth edges. We again evaluate accuracy using the same three standard metrics: ODS, OIS and AP.

Can an image edge detector be trained using motion edges? Our first experiment tests this question. We use all \emph{ground truth} motion edges available in VSB ($591$ images) to train both \alg{se} and \alg{he}. The results are reported in Table~\ref{table:bsds} (\alg{se-vsb}, \alg{he-vsb}). For both methods, results are within 2-4 points ODS compared to training with image edge supervision (\alg{se-bsds}, \alg{he-bsds}). Our results suggest that using motion edges to learn an image edge detector is feasible.

\begin{table}\centering\footnotesize
\begin{tabular}{lr}
\hspace{-2mm}
\includegraphics[width=.21\textwidth]{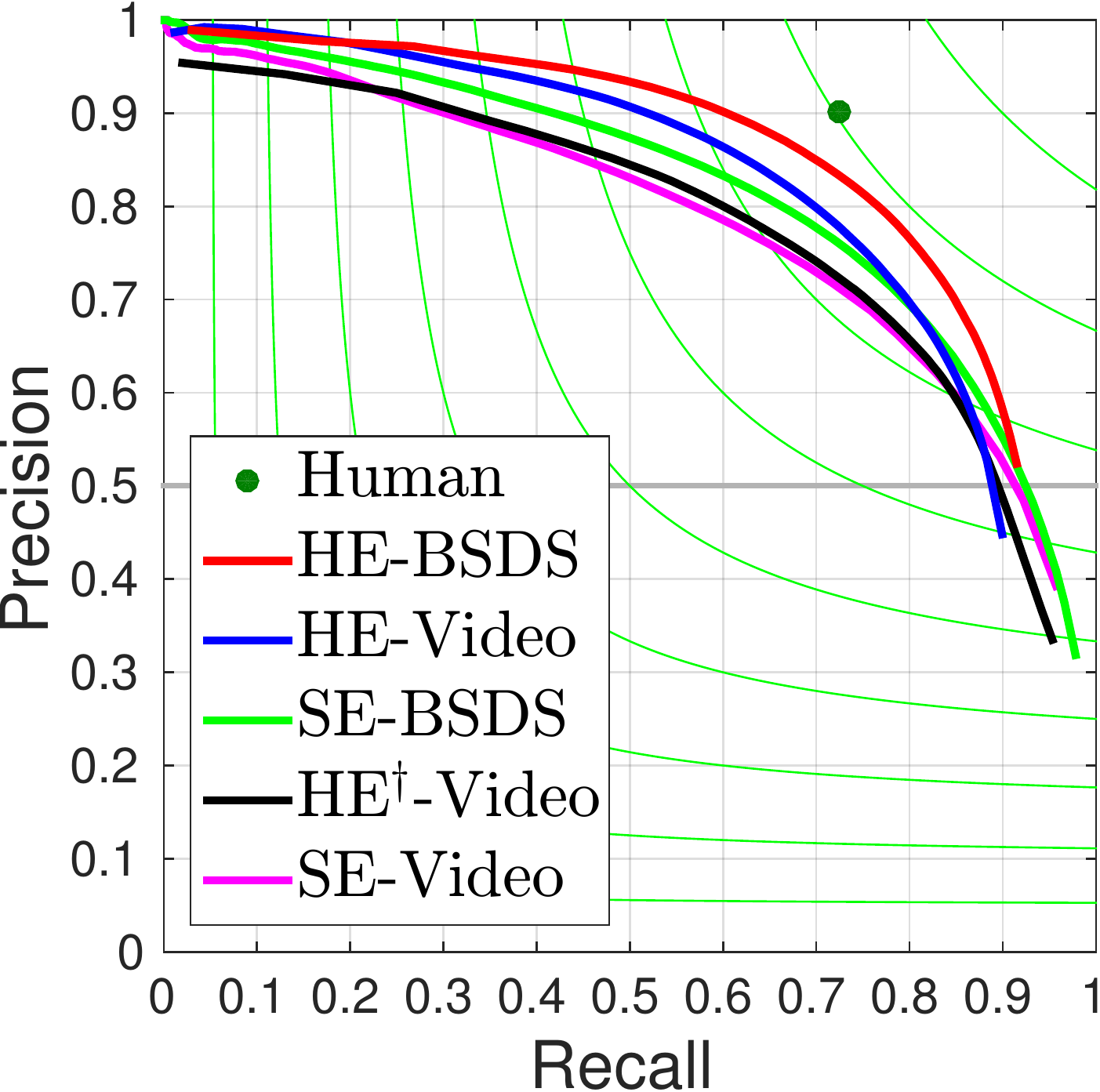} \hspace{-2mm} &
\begin{tabular}[b]{@{}l|ccc@{}}
Method & ODS &  OIS & AP \\
 \hline
 \alg{human} & .80 & .80 & - \\
 \hline
 \alg{se-bsds} & .746 & .767 & .803 \\
 \alg{se-vsb} & .722 & .744 & .757 \\
 \alg{se-video} & .724 & .748 & .763 \\
 \hline
 \alg{he-bsds} & .785 & .803 & .791 \\
 \alg{he-vsb} & .745 & .772 & .769 \\
 \alg{he-video} & .748 & .770 & .772 \\
 \hline
 \alg{he$^\dagger$-bsds} & .760 & .774 & .790 \\
 \alg{he$^\dagger$-vsb} & .719 & .735 & .751 \\
 \alg{he$^\dagger$-video} & .726 & .745 & .761 \\
\end{tabular}
\end{tabular}
\vspace{2mm}
\caption{Edge detection results on BSDS test set. We report results for \alg{se} and \alg{he} using three training scenarios: \alg{bsds}, \alg{vsb}, and \alg{video} (unsupervised). \alg{he} uses the VGG network pre-trained on ImageNet, \alg{he$^\dagger$} indicates that network is trained from scratch.}
\label{table:bsds}
\end{table}

We next present results using videos as the supervisory signal (\alg{se-video}, \alg{he-video}). \alg{se-video} achieves an ODS of $.724$ compared to $.746$ for the supervised case (\alg{se-bsds}). Results for \alg{he} are similar ($.748$ versus $.785$). As these results show, using video supervision achieves competitive results (within 3-5\%). Interestingly, learning from video slightly outperforms training using the ground truth motion edges. We attribute this to the small size of VSB.

For \alg{he}, we experiment with starting with an ImageNet pre-trained model (\alg{he}) and training from scratch (\alg{he$^\dagger$}). Pre-training on ImageNet benefits \alg{he} across all training scenarios. This is encouraging as it implies that object level knowledge is useful for edge detection. On the other hand, our video supervision scheme also benefits from ImageNet pre-training, thus implying that in our current setup we are not training the model to its full potential.

\begin{figure}\centering
\includegraphics[width=.235\textwidth]{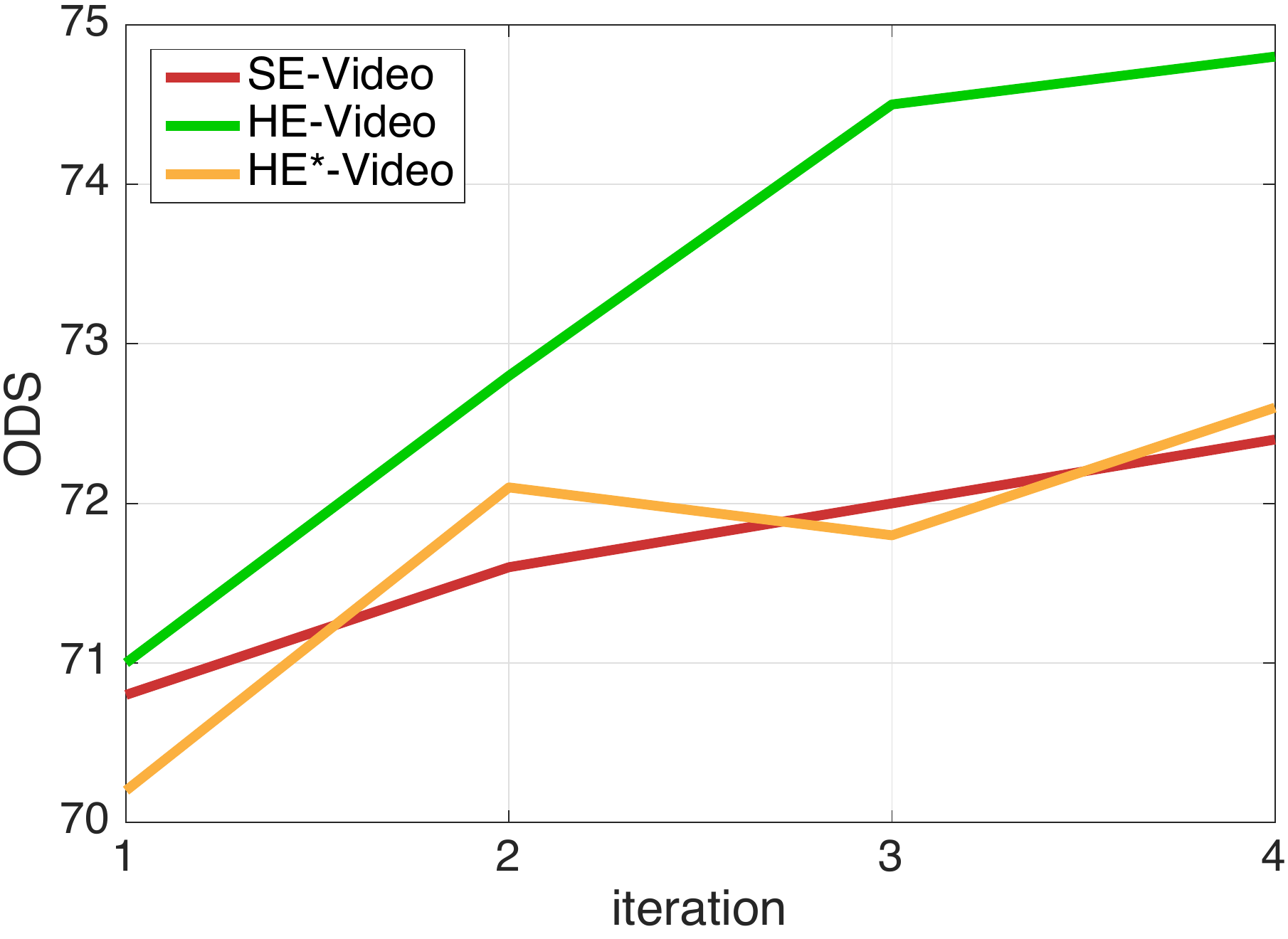}
\includegraphics[width=.235\textwidth]{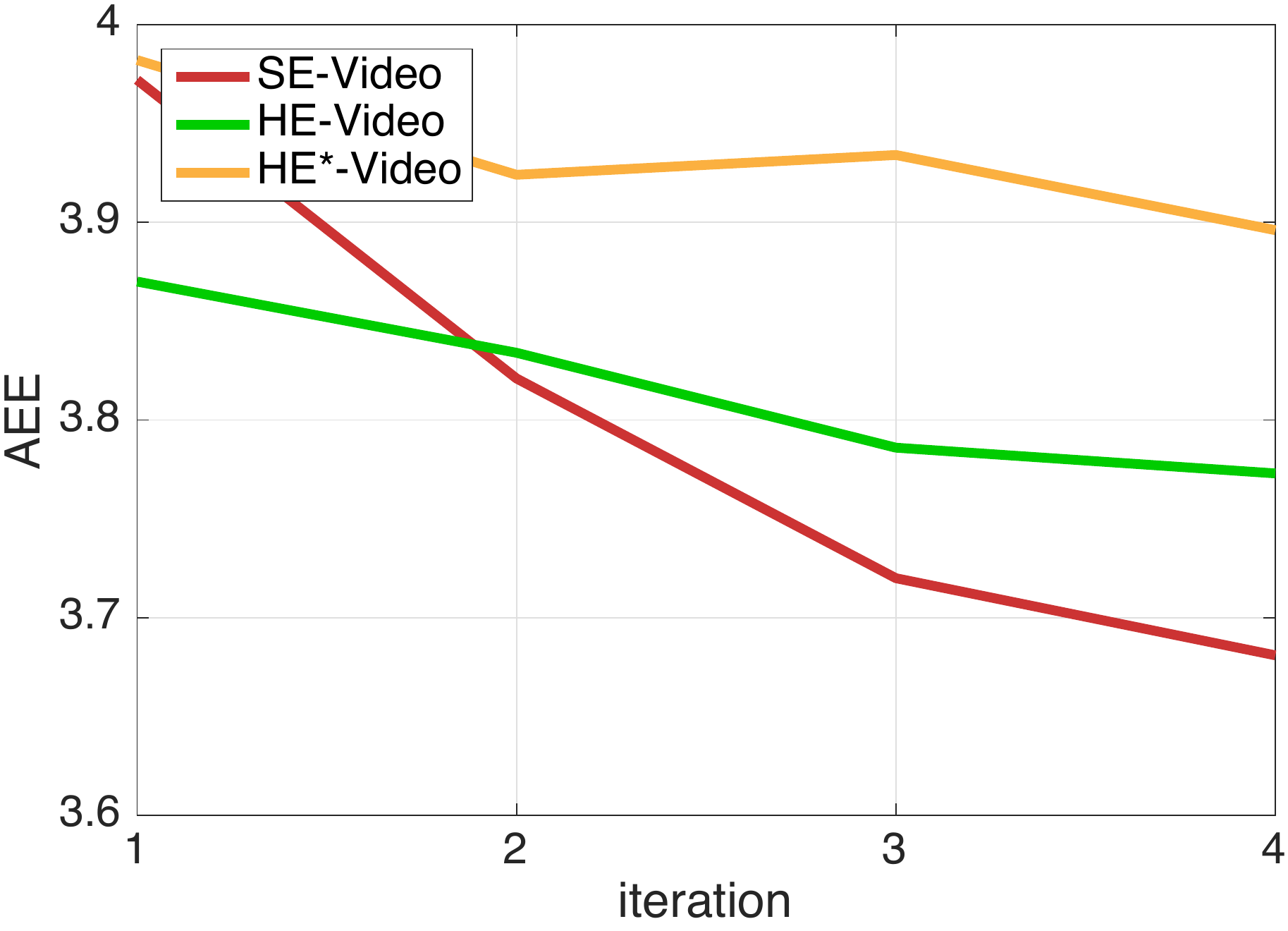}
\caption{Convergence of ODS and AEE over iterations. See text.}
\label{fig:iteration}
\end{figure}

To probe how performance evolves, we plot the ODS scores at each iteration for both methods in Figure~\ref{fig:iteration}. Raw image gradient at iteration $0$ has ODS of $.543$ (not shown). Our iterative process provides a significant improvement from the image gradient, with most of the gains coming in the first iteration. Performance saturates after about 4 iterations (for the last iteration, we use 4 million samples for \alg{se} and $80$ epochs for \alg{he}, which slightly increases accuracy).

We provide visualizations of edge results (before NMS) in Figure~\ref{fig:qualitative}. \alg{se} misses some weak edges but edges are well aligned to the image content. \alg{he} generally produces thicker edges due to use of downsampled conv feature maps which makes it difficult to produce sharp image edges. \alg{he-video}/\alg{he$^\dagger$-video} results have thinner edges than \alg{he-bsds}/\alg{he$^\dagger$-bsds}, potentially due to the sampling strategy used for training with motion edges. When training using video, we also observe that the edge detection output is less well localized and more likely to miss weak edges, which likely accounts for much of the performance differences.

\subsection{Optical Flow}

We benchmark optical flow results on the Middlebury~\cite{baker2011database} and MPI Sintel~\cite{butler2012sintel} datasets. Middlebury is widely used and consists of complex motions with small displacements. Sintel is obtained from animated sequences and features large displacements and challenging lighting conditions. We use the `final' version of Sintel  and test on the train set with public ground truth. As our goal is to test the quality of generated edge maps, we focus only on variants of EpicFlow~\cite{revaud2015epic}, the highest performing method on Sintel as of CVPR 2015.

Table~\ref{table:flow} shows the average endpoint error (AEE) of EpicFlow using different edge maps for Sintel and Middleburry. Most edge maps give rise to relatively similar results (AEE around $3.6\app3.8$) on Sintel. In particular, originally EpicFlow used \alg{se-bsds} edges; the results with \alg{se-video} edges are nearly identical. Top results are obtained with \alg{he-bsds}, while \alg{he-video} and \alg{he$^\dagger$-video} are slightly worse. On Middleburry the method rankings are similar.

As an upper bound, we also include EpicFlow given ground truth (\alg{GT}) motion edges (derived from the ground truth flow). Accuracy is only slightly better than with the best learned edge maps.
This suggests that the performance of EpicFlow is saturated given the current matches.

Finally, in Figure~\ref{fig:iteration} we plot AEE on Sintel for each iteration. All methods improve over the initial flow (AEE $4.016$, not shown) and results again saturate after a few iterations.

\begin{table}\centering\small
\begin{tabular}{l|cc}
Contour & MPI-Sintel & Middlebury \\
 \hline
 \alg{gt} motion edges    & 3.588 & - \\
 \hline
 \alg{se-bsds}            & 3.686 & .380 \\
 \alg{se-video}           & 3.681 & .385 \\
 \hline
 \alg{he-bsds} 		     & 3.608 & .298 \\
 \alg{he-video}           & 3.773 & .308\\
 \alg{he$^\dagger$-video} & 3.896 & .390 \\
\end{tabular}\vspace{2mm}
\caption{Accuracy of EpicFlow with various edge maps (AEE).}
\label{table:flow}
\end{table}

\subsection{Object Detection}

Finally, we test whether our unsupervised training scheme for edge detection can be used to pre-train a network for object detection. The question of whether strong supervision is necessary for learning a good visual representation for object detection is of much recent interest~\cite{wang2015unsupervised, doersch2015unsupervised}. While not the focus of our work, we demonstrate that our scheme can likewise be used for network initialization.

For these experiments, we use the \alg{he$^\dagger$} edge detector (without ImageNet pre-training). We perform experiments using PASCAL VOC 2007~\cite{everingham2014pascal} and the Fast R-CNN~\cite{ross2015frcnn} object detector with proposals from~\cite{Uijlings13IJCV}. Results are evaluated by mean Average Precision (mAP). We compare results using two networks, VGG~\cite{Simonyan14vgg} and ZF~\cite{zeiler2014visualizing}, and four pre-training schemes: ImageNet pre-training; no pre-training; pre-trained on BSDS (\alg{he$^\dagger$-bsds}), and pre-trained using video (\alg{he$^\dagger$-video}). All networks are fine-tuned using the train-val set for $40K$ iterations ($120K$ iterations when training from scratch). Results are summarized in Table~\ref{table:voc_pretrain}.

\textbf{VGG Results:} We attempted to train VGG~\cite{Simonyan14vgg} from scratch on VOC (with various learning rates plus batch normalization and dropout) but failed to obtain meaningful results. Even after $120K$ iterations detection performance remains poor ($\app15$ mAP). When the network is pre-trained on BSDS for edge detection, we were able to achieve $42.1$ mAP on PASCAL. Interestingly, when training using video, we observe a further $2$ point improvement in mAP (even though the same network is inferior for edge detection).

\textbf{ZF Results:} We also experimented with training a smaller ZF network~\cite{zeiler2014visualizing} which has only $5$ convolutional layers. We modify the network slightly for edge detection to facilitate the alignment between outputs from different layers\footnote{We change kernel size of all pooling layers to $2$ and modify padding to $3$ and $2$ in conv1 and conv2, respectively. We also attach classifiers ($1\times 1$ convs) on all conv layers and up-sample and merge the results into a single edge map as in~\cite{Xie2015hed}. Finally, when training from scratch, we add batch normalization after every conv layer. The ZF network has an ODS of $.715$ when trained using BSDS and $.681$ when trained using videos.}. With ImageNet pre-training, Fast R-CNN with our modified ZF network achieves $58.6$ mAP on PASCAL, which is close to the ZF result originally reported in~\cite{ross2015frcnn}. With no pre-training, mAP drops to $38.2$. Pre-trained for edge detection, either with or without supervision, improves results by \app3 mAP over training from scratch.

Overall we conclude that pre-training for edge detection (with or without supervision) improves the performance of training an object detector from scratch. However, ImageNet pre-training still achieves substantially better results.

\begin{table}\centering\small
\begin{tabular}{l|c}
  pre-training & mAP \\
  \hline
  \alg{ImageNet} & 66.9 \\
  \alg{none} & 15.6 \\
  \alg{he$^\dagger$-bsds} & 42.1 \\
  \alg{he$^\dagger$-video} & 44.2 \\
\end{tabular}\hspace{10mm}
\begin{tabular}{l|c}
  pre-training & mAP \\
  \hline
  \alg{ImageNet} & 58.6 \\
  \alg{none} & 38.2 \\
  \alg{he$^\dagger$-bsds} & 41.4 \\
  \alg{he$^\dagger$-video} & 41.1 \\
\end{tabular}\vspace{2mm}
\caption{Object detection results (mean AP) on PASCAL VOC 2007 test using VGG (left) and ZF (right). See text for details.}
\label{table:voc_pretrain}
\end{table}

\subsection{Limitations}

\emph{Why doesn't unsupervised training outperform supervised training for edge detection?} In theory, a sufficiently large corpus of video should provide an unlimited training set and an edge detector trained on this massive corpus should outperform the much smaller supervised training set. However, there are a number of issues that currently limit performance. (1)~Existing flow methods are less accurate at weak image edges, in addition, our alignment scheme also removes weak edges. Thus weak image edges are missing from our training set. (2)~Further improving image edges does not  improve optical flow, see Table~\ref{table:motion}. We conjecture that the matches between frames are the limiting factor for EpicFlow and until these are improved neither optical flow nor edges will improve in the current scheme. (3)~Training is harmed by noisy labels, and in particular, the missing positive labels. The false negatives, if not handled properly, tend to dominate the gradients in the late stages of training.

\emph{Is the unsupervised learning scheme capturing object-level information?} The extent of an object is defined by its edges and conversely many edges can only be identified with knowledge of objects. Our results on both edge and object detection support this connection: on one hand, ImageNet pre-training is useful for edge detection, possibly because it injects object-level information into the network. On the other hand, pre-training a network for edge detection also improves object detection. In principle, an edge detection network has to learn high-level shape information, which might explain the effectiveness of pre-training. However, we observe that pre-training on ImageNet still benefits edge detection under all scenarios; moreover, ImageNet pre-training is still substantially better than video pre-training for object detection. Hence, perhaps unsurprisingly, the current unsupervised scheme is not as effective as ImageNet pre-training at capturing object level information.

\newcommand{\bsdsFigHelper}[2]{
  \hspace{-2mm}\rotatebox{90}{\begin{minipage}{
    .126\linewidth}\center#1\end{minipage}} &
  \includegraphics[width=.187\linewidth]{figures/results/157087#2} &
  \includegraphics[width=.187\linewidth]{figures/results/97010#2} &
  \includegraphics[width=.187\linewidth]{figures/results/196062#2} &
  \includegraphics[width=.187\linewidth]{figures/results/69000#2} &
  \includegraphics[width=.187\linewidth]{figures/results/81066#2}\\
}
\begin{figure*}\center
\begin{tabular}{r@{\hskip 1mm}c@{\hskip 1mm}c@{\hskip 1mm}c@{\hskip 1mm}c@{\hskip 1mm}c}
  \bsdsFigHelper{}{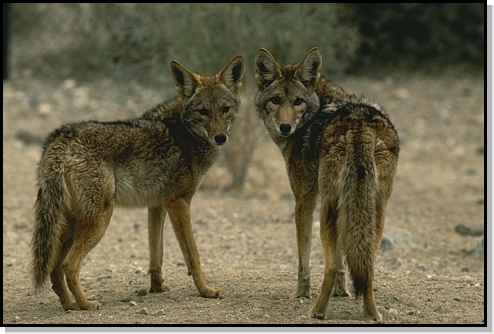}
  \bsdsFigHelper{\alg{human}}{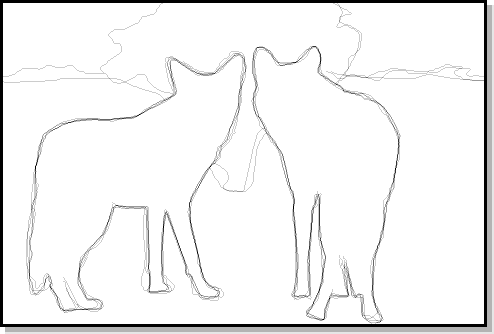}
  \bsdsFigHelper{\alg{se-bsds}}{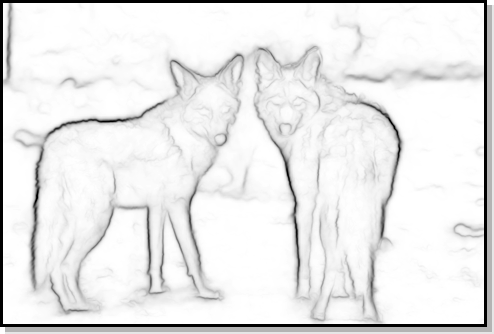}
  \bsdsFigHelper{\alg{se-video}}{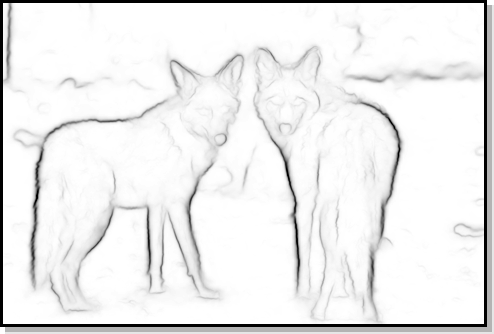}
  \bsdsFigHelper{\alg{he-bsds}}{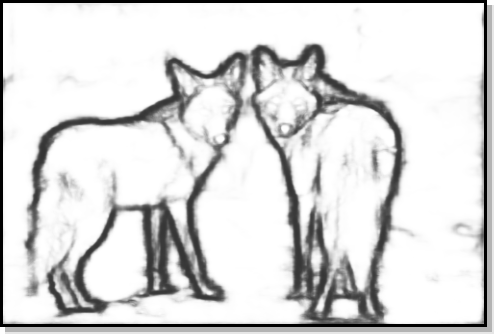}
  \bsdsFigHelper{\alg{he-video}}{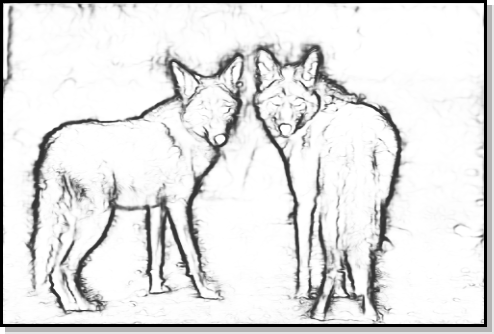}
  \bsdsFigHelper{\alg{he$^\dagger$-bsds}}{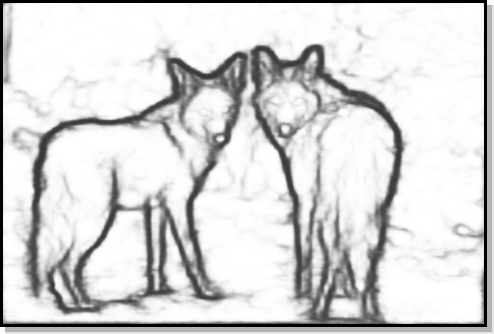}
  \bsdsFigHelper{\alg{he$^\dagger$-video}}{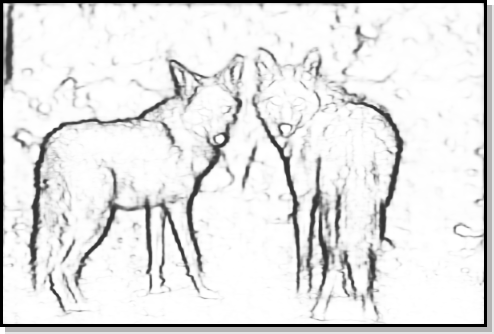}
\end{tabular}
\caption{Illustration of edge detection results on five sample images (same as used in~\cite{Dollar2015fast}). The first two rows show the original image and ground truth. The second and third rows are results from \alg{se}, trained using \alg{bsds} or \alg{video}. The remaining rows show the results of variants of \alg{he} on \alg{bsds} or \alg{video}. \alg{he$^\dagger$} indicates that the network is trained from scratch. Use viewer zoom functionality to see fine details.}\label{fig:qualitative}
\end{figure*}

\section{Discussion}

In this work, we proposed to harvest motion edges to learn an edge detector from video without explicit supervision. We developed an iterative process that alternated between updating optical flow using edge results, and learning an edge detector based on the flow fields, leading to increasingly accurate edges and flows.

The main result of our paper is that edge detectors, trained using our unsupervised scheme, approach the same level of performance as fully supervised training.

We additionally demonstrated our approach can serve a novel unsupervised pre-training scheme for deep networks. While the gains from pre-training were modest, we believe this is a promising direction for future exploration.

In the long run we believe that unsupervised learning of edge detectors has the potential to outperform supervised training as the unsupervised approach has access to unlimited data. Our work is the first serious step in this direction.\\
\\
\textbf{Acknowledgements} \small We thank Saining Xie for help with the \alg{he} detector and Ahmad Humayun, Yan Zhu, and Yuandong Tian and many others for valuable discussions and feedback. The work was conducted when Yin Li was an intern at FAIR. Yin Li gratefully acknowledges the support of the Intel ISTC-PC while completing the writing of the paper at Georgia Tech.

\newpage
{\small\bibliographystyle{ieee}\bibliography{biblio}}

\begin{thebibliography}{10}\itemsep=-1pt

\bibitem{achanta2012slic}
R.~Achanta, A.~Shaji, K.~Smith, A.~Lucchi, P.~Fua, and S.~Susstrunk.
\newblock {SLIC} superpixels compared to state-of-the-art superpixel methods.
\newblock {\em PAMI}, 2012.

\bibitem{Arbelaez2011PAMI}
P.~Arbelaez, M.~Maire, C.~Fowlkes, and J.~Malik.
\newblock Contour detection and hierarchical image segmentation.
\newblock {\em PAMI}, 2011.

\bibitem{Arbelaez2014CVPR}
P.~Arbel{\'a}ez, J.~Pont-Tuset, J.~T. Barron, F.~Marques, and J.~Malik.
\newblock Multiscale combinatorial grouping.
\newblock In {\em CVPR}, 2014.

\bibitem{baker2011database}
S.~Baker, D.~Scharstein, J.~Lewis, S.~Roth, M.~J. Black, and R.~Szeliski.
\newblock A database and evaluation methodology for optical flow.
\newblock {\em IJCV}, 2011.

\bibitem{bertasuis2015high}
G.~Bertasius, J.~Shi, and L.~Torresani.
\newblock High-for-low and low-for-high: Efficient boundary detection from deep
  object feat.~and its app.~to high-level vision.
\newblock In {\em ICCV}, 2015.

\bibitem{brox2011large}
T.~Brox and J.~Malik.
\newblock Large displacement optical flow: descriptor matching in variational
  motion estimation.
\newblock {\em PAMI}, 2011.

\bibitem{butler2012sintel}
D.~J. Butler, J.~Wulff, G.~B. Stanley, and M.~J. Black.
\newblock A naturalistic open source movie for optical flow evaluation.
\newblock In {\em ECCV}, 2012.

\bibitem{Canny1986PAMI}
J.~Canny.
\newblock A computational approach to edge detection.
\newblock {\em PAMI}, 1986.

\bibitem{doersch2015unsupervised}
C.~Doersch, A.~Gupta, and A.~A. Efros.
\newblock Unsupervised visual representation learning by context prediction.
\newblock In {\em ICCV}, 2015.

\bibitem{Dollar2006CVPR}
P.~Doll\'ar, Z.~Tu, and S.~Belongie.
\newblock Supervised learning of edges and object boundaries.
\newblock In {\em CVPR}, 2006.

\bibitem{Dollar2015fast}
P.~Doll{\'a}r and C.~L. Zitnick.
\newblock Fast edge detection using structured forests.
\newblock {\em PAMI}, 2015.

\bibitem{everingham2014pascal}
M.~Everingham, S.~A. Eslami, L.~Van~Gool, C.~K. Williams, J.~Winn, and
  A.~Zisserman.
\newblock The pascal visual object classes challenge: A retrospective.
\newblock {\em IJCV}, 2014.

\bibitem{Ferrari2008PAMI}
V.~Ferrari, L.~Fevrier, F.~Jurie, and C.~Schmid.
\newblock Groups of adjacent contour segments for object detection.
\newblock {\em PAMI}, 2008.

\bibitem{Fram1975IEEE}
J.~R. Fram and E.~S. Deutsch.
\newblock On the quantitative evaluation of edge detection schemes and their
  comparison with human performance.
\newblock {\em IEEE TOC}, 1975.

\bibitem{Freeman91PAMI}
W.~T. Freeman and E.~H. Adelson.
\newblock The design and use of steerable filters.
\newblock {\em PAMI}, 1991.

\bibitem{galasso2013unified}
F.~Galasso, N.~S. Nagaraja, T.~Jimenez~Cardenas, T.~Brox, and B.~Schiele.
\newblock A unified video segmentation benchmark: Annotation, metrics and
  analysis.
\newblock In {\em ICCV}, 2013.

\bibitem{ross2015frcnn}
R.~Girshick.
\newblock Fast {R-CNN}.
\newblock In {\em ICCV}, 2015.

\bibitem{horn1981determining}
B.~K. Horn and B.~G. Schunck.
\newblock Determining optical flow.
\newblock In {\em 1981 Technical symposium east}. International Society for
  Optics and Photonics, 1981.

\bibitem{ioffe2009batch}
S.~Ioffe and C.~Szegedy.
\newblock Batch normalization: Accelerating deep network training by reducing
  internal covariate shift.
\newblock In {\em ICML}, 2015.

\bibitem{Kivinen2014AISTATS}
J.~J. Kivinen, C.~K. Williams, and N.~Heess.
\newblock Visual boundary prediction: A deep neural prediction network and
  quality dissection.
\newblock In {\em AISTATS}, 2014.

\bibitem{koffka2013principles}
K.~Koffka.
\newblock {\em Principles of Gestalt psychology}.
\newblock Routledge, 2013.

\bibitem{lee2014deeply}
C.-Y. Lee, S.~Xie, P.~Gallagher, Z.~Zhang, and Z.~Tu.
\newblock Deeply-supervised nets.
\newblock In {\em AISTATS}, 2015.

\bibitem{Lim2013CVPR}
J.~Lim, C.~L. Zitnick, and P.~Doll\'ar.
\newblock Sketch tokens: A learned mid-level representation for contour and
  object detection.
\newblock In {\em CVPR}, 2013.

\bibitem{Lucas1981iterative}
B.~D. Lucas, T.~Kanade, et~al.
\newblock An iterative image registration technique with an application to
  stereo vision.
\newblock In {\em IJCAI}, 1981.

\bibitem{Martin2004PAMI}
D.~Martin, C.~Fowlkes, and J.~Malik.
\newblock Learning to detect natural image boundaries using local brightness,
  color, and texture cues.
\newblock {\em PAMI}, 2004.

\bibitem{mobahi2009deep}
H.~Mobahi, R.~Collobert, and J.~Weston.
\newblock Deep learning from temporal coherence in video.
\newblock In {\em ICML}, 2009.

\bibitem{ostrovsky2009visual}
Y.~Ostrovsky, E.~Meyers, S.~Ganesh, U.~Mathur, and P.~Sinha.
\newblock Visual parsing after recovery from blindness.
\newblock {\em Psychological Science}, 2009.

\bibitem{Prest2012CVPR}
A.~Prest, C.~Leistner, J.~Civera, C.~Schmid, and V.~Ferrari.
\newblock Learning object class detectors from weakly annotated video.
\newblock In {\em CVPR}, 2012.

\bibitem{ranzato2014video}
M.~Ranzato, A.~Szlam, J.~Bruna, M.~Mathieu, R.~Collobert, and S.~Chopra.
\newblock Video (language) modeling: a baseline for generative models of
  natural videos.
\newblock In {\em ICLR}, 2015.

\bibitem{Ren2012NIPS}
X.~Ren and B.~Liefeng.
\newblock Discriminatively trained sparse code gradients for contour detection.
\newblock In {\em NIPS}, 2012.

\bibitem{revaud2015epic}
J.~Revaud, P.~Weinzaepfel, Z.~Harchaoui, and C.~Schmid.
\newblock {EpicFlow: Edge-Preserving Interpolation of Correspondences for
  Optical Flow}.
\newblock In {\em CVPR}, 2015.

\bibitem{rublee2011orb}
E.~Rublee, V.~Rabaud, K.~Konolige, and G.~Bradski.
\newblock {ORB}: an efficient alternative to {SIFT} or {SURF}.
\newblock In {\em ICCV}, 2011.

\bibitem{ILSVRC15}
O.~Russakovsky, J.~Deng, H.~Su, J.~Krause, S.~Satheesh, S.~Ma, Z.~Huang,
  A.~Karpathy, A.~Khosla, M.~Bernstein, A.~C. Berg, and L.~Fei-Fei.
\newblock {ImageNet Large Scale Visual Recognition Challenge}.
\newblock {\em IJCV}, 2015.

\bibitem{Simonyan14vgg}
K.~Simonyan and A.~Zisserman.
\newblock Very deep convolutional networks for large-scale image recognition.
\newblock {\em arXiv:1409.1556}, 2014.

\bibitem{sironi2014multiscale}
A.~Sironi, V.~Lepetit, and P.~Fua.
\newblock Multiscale centerline detection by learning a scale-space distance
  transform.
\newblock In {\em CVPR}, 2014.

\bibitem{srivastava2015unsupervised}
N.~Srivastava, E.~Mansimov, and R.~Salakhutdinov.
\newblock Unsupervised learning of video representations using {LSTMs}.
\newblock In {\em ICML}, 2015.

\bibitem{szegedy2014scalable}
C.~Szegedy, S.~Reed, D.~Erhan, and D.~Anguelov.
\newblock Scalable, high-quality object detection.
\newblock {\em arXiv:1412.1441}, 2014.

\bibitem{taylor2010convolutional}
G.~Taylor, R.~Fergus, Y.~LeCun, and C.~Bregler.
\newblock Convolutional learning of spatio-temporal feat.
\newblock In {\em ECCV}, 2010.

\bibitem{Uijlings13IJCV}
J.~Uijlings, K.~van~de Sande, T.~Gevers, and A.~Smeulders.
\newblock Selective search for object recog.
\newblock {\em IJCV}, 2013.

\bibitem{UllmanPAMI91}
S.~Ullman and R.~Basri.
\newblock Recognition by linear combinations of models.
\newblock {\em PAMI}, 1991.

\bibitem{wang2015unsupervised}
X.~Wang and A.~Gupta.
\newblock Unsupervised learning of visual representations using videos.
\newblock In {\em ICCV}, 2015.

\bibitem{weinzaepfel2013dm}
P.~Weinzaepfel, J.~Revaud, Z.~Harchaoui, and C.~Schmid.
\newblock {DeepFlow: Large displacement optical flow with deep matching}.
\newblock In {\em ICCV}, 2013.

\bibitem{weinzaepfel2015learning}
P.~Weinzaepfel, J.~Revaud, Z.~Harchaoui, and C.~Schmid.
\newblock {Learning to Detect Motion Boundaries}.
\newblock In {\em CVPR}, 2015.

\bibitem{Xie2015hed}
S.~Xie and Z.~Tu.
\newblock Holistically-nested edge detection.
\newblock In {\em ICCV}, 2015.

\bibitem{zeiler2014visualizing}
M.~D. Zeiler and R.~Fergus.
\newblock Visualizing and understanding convolutional networks.
\newblock In {\em ECCV}, 2014.

\bibitem{Zitnick2014ECCV}
C.~L. Zitnick and P.~Doll{\'{a}}r.
\newblock Edge boxes: Locating object proposals from edges.
\newblock In {\em ECCV}, 2014.

\bibitem{Zitnick2012CVPR}
C.~L. Zitnick and D.~Parikh.
\newblock The role of image understanding in contour detection.
\newblock In {\em CVPR}, 2012.

\end{thebibliography}
\end{document}